\documentclass[sigconf]{acmart}

\usepackage[labelfont=normalfont,textfont=normalfont]{caption}
\usepackage{graphicx} % Required for inserting images

\usepackage{algorithm}
\usepackage{algorithmic}
\usepackage{graphicx}
\usepackage{float}

\AtBeginDocument{%
  }

\begin{document}

\title{Deepmechanics}

\author{Abhay Shinde}
\email{abhay3927327@mccmulund.ac.in}
\affiliation{%
  \institution{Deep Forest Sciences}
  \city{Mumbai}
  \state{Maharashtra}
  \country{India}
}
\author{Aryan Amit Barsainyan}
\email{aryan@deepforestsci.com}
\affiliation{%
  \institution{Deep Forest Sciences}
  \city{Bengaluru}
  \state{Karnataka}
  \country{India}
}
\author{Jose Siguenza}
\email{jose@deepforestsci.com}
\affiliation{%
  \institution{Deep Forest Sciences}
  \city{Palo alto}
  \state{California}
  \country{USA}
}
\author{Ankita Vaishnobi Bisoi}
\email{f20212306@goa.bits-pilani.ac.in}
\affiliation{%
  \institution{Deep Forest Sciences}
  \city{Dabolim}
  \state{Goa}
  \country{India}
}
\author{Rakshit Kr. Singh}
\email{rakshit@deepforestsci.com}
\affiliation{%
  \institution{Deep Forest Sciences}
  \city{Bengaluru}
  \state{Karnataka}
  \country{India}
}
\author{Bharath Ramsundar}
\email{bharath@deepforestsci.com}
\affiliation{%
  \institution{Deep Forest Sciences}
  \city{Palo alto}
  \state{California}
  \country{USA}
}

\begin{abstract}
Physics-informed deep learning models have emerged as powerful tools for learning dynamical systems. These models directly encode physical principles into network architectures. However, systematic benchmarking of these approaches across diverse physical phenomena remains limited, particularly in conservative and dissipative systems. In addition, benchmarking that has been done thus far does not integrate out full trajectories to check stability. In this work, we benchmark three prominent physics-informed architectures such as Hamiltonian Neural Networks (HNN), Lagrangian Neural Networks (LNN), and Symplectic Recurrent Neural Networks (SRNN) using the DeepChem framework, an open-source scientific machine learning library. We evaluate these models on six dynamical systems spanning classical conservative mechanics (mass-spring system, simple pendulum, double pendulum, and three-body problem, spring-pendulum) and non-conservative systems with contact (bouncing ball). We evaluate models by computing error on predicted trajectories and evaluate error both quantitatively and qualitatively. We find that all benchmarked models struggle to maintain stability for chaotic or nonconservative systems. Our results suggest that more research is needed for physics-informed deep learning models to learn robust models of classical mechanical systems. 
\end{abstract}



\maketitle

\section{Introduction}
Modeling and predicting the dynamics of physical systems remains a fundamental challenge in robotics, control theory, and computational physics. Traditional approaches rely on manually designed differential equations based on first principles, such as Newtonian mechanics, Lagrangian dynamics, or Hamiltonian systems. While these approaches are very effective, they have significant limitations when applied to model complex real-world systems, particularly those involving contact ~\citep{le2024contact}, friction, and energy dissipation. In robotics, where systems interact with their environment through grasping, manipulation, and locomotion, accurately modeling these effects is difficult and often contributes to sim-to-real gaps ~\citep{aljalbout2025reality}.  

Recent advances in physics-informed deep learning have revealed new possibilities for dynamics recovery from data while satisfying the basic principles of physics. Methods such as Hamiltonian Neural Networks (HNNs)~\citep{greydanus2019hamiltonian}, Lagrangian Neural Networks (LNNs)~\citep{cranmer2020lagrangian}, and Symplectic Recurrent Neural Networks (SRNNs)~\citep{chen2019symplectic} have been shown to possess the capability to learn dynamics that conserve energy better than baseline neural networks with improved generalization and long-term behavior. These methods incorporate the physical principle of energy conservation and symplectic structure.

However, there is a significant gap in the critical assessment of the performance of these physics-informed architectures on various dynamical systems, particularly when they are challenged by non-conservative systems such as damping and contact. Furthermore, the lack of standardized implementations and benchmarking tools makes it difficult to replicate results and to reliably compare the performance of different physics-informed neural network approaches.

\begin{figure*}
  \centering
  \includegraphics[width=\textwidth]{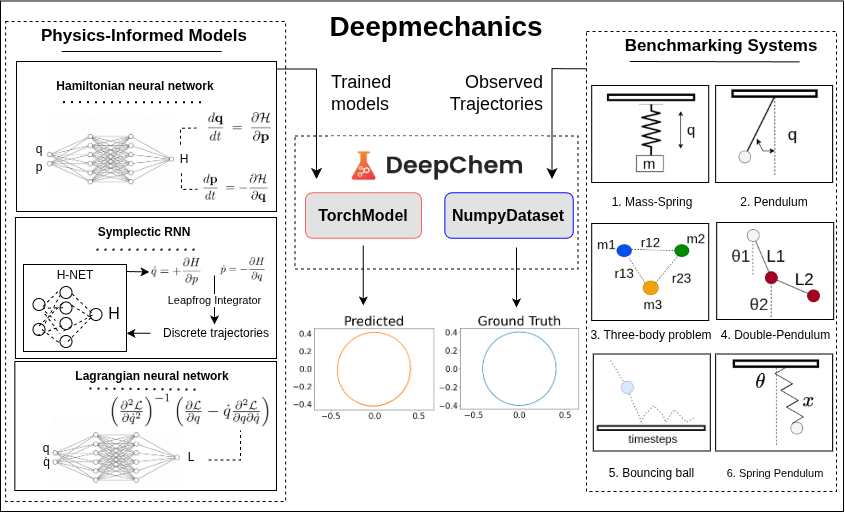}
  \caption{Deepmechanics framework.
  Left : Three physics-informed models (HNN, LNN, SRNN) implemented in Deepchem.
  Middle : Deepchem infrastructure as TorchModel and NumpyDataset.
  Right : Benchmarked this implemented models in six different systems .
  }
    \label{fig:deepmechanics}
\end{figure*}

In this work, we present \textit{DeepMechanics} (Figure~\ref{fig:deepmechanics}), a systematic benchmark of physics-informed neural networks implemented within the DeepChem~\citep{ramsundar2018molecular}, which is an open-source Python library that was initially developed for machine learning tasks in drug discovery and materials science ~\citep{singh2025chemberta}~\citep{ahmad2022chemberta} ~\citep{chithrananda2020chemberta}. DeepChem provides high-level abstractions like NumpyDataset and TorchModel to handle data and train models. We extend DeepChem with implementations of HNN, LNN, and SRNN models and perform a thorough benchmark on six dynamical systems: two classical conservative systems (mass-spring, simple pendulum), three chaotic systems (double pendulum, three-body problem, spring pendulum), and one non-conservative system with contact (bouncing ball).

\section{Background}

\subsection{Physics-informed deep learning}
Physics-informed neural networks integrate physical constraints into neural network architectures to improve the modeling of dynamical systems. Research in this area has explored various strategies, from solving differential equations with neural networks~\cite{lagaris1998artificial}. to designing architectures that preserve specific physical properties like energy conservation and symplectic geometry. In this work we focus on three prominent architectures: Hamiltonian Neural Networks (HNNs)~\citep{greydanus2019hamiltonian}, Lagrangian Neural Networks (LNNs)~\citep{cranmer2020lagrangian}, and Symplectic Recurrent Neural Networks (SRNNs)~\citep{chen2019symplectic}.

\subsection{Energy based models}
Many physical systems can be described through an energy function. Hamiltonian Neural Networks (HNNs)~\citep{greydanus2019hamiltonian} and Lagrangian Neural Networks (LNNs)~\citep{cranmer2020lagrangian} parameterize energy functions based on Hamiltonian dynamics and Euler-Lagrange equations respectively. These models compute gradients of the learned energy function with respect to state variables (position, momentum, velocity) to derive equations of motion, which are then integrated using ODE solvers to predict future system states.

\subsection{Structure preserving models}
Numerical integration techniques such as ODEs and PDEs introduce discretization errors when the hamiltonian systems gets complex such as the three body problem. These errors accumulate over time, causing trajectories to diverge quickly which prevents the conservation of energy. Structure preserving models like Symplectic-RNN (SRNNs)~\citep{chen2019symplectic} tackle this problem by parameterizing the hamiltonian to learn discrete trajectories using leapfrog integrator which gets backpropogated over the time steps.

\subsection{Contact in Dynamical Systems}
While energy-preserving models perform well on smooth trajectories, they struggle to accurately capture systems involving contact and friction. Past work has extended HNNs and LNNs with differentiable contact model~\citep{zhong2021extending}. In this work, we evaluate HNNs, LNNs, and SRNNs on the bouncing ball system to analyze contact dynamics.

\section{Experiments and Results}
For each system, models are trained on full observed trajectories to learn complete system behavior. Evaluation is performed with initial state values from the test split of the generated trajectories and produces predicted trajectories via numerical integration with model-specific schemes: SciPy ~\citep{2020SciPy-NMeth} routines such as \texttt{solve\_ivp} (RK45) and \texttt{odeint} for HNN and LNN, respectively, and DeepChem’s leapfrog integrator for SRNN. Model performance is assessed through quantitative metrics such as mean squared error (MSE), mean absolute error (MAE), root mean squared error (RMSE), and the standard deviation (STD) and variance (VAR), computed between observed and predicted trajectories, and qualitative analysis of phase-space plots and state variable evolution to evaluate preservation of physical properties such as energy conservation and trajectory stability.

\subsection{Mass-Spring System}
A mass-spring system models a point mass $m$ attached to a spring with stiffness $k$ which moves along a single dimension.

\subsubsection{\textbf{Hamiltonian Neural Network (HNN)}}
The Hamiltonian equation for mass-spring is given by~\citep{mitofsky_mass_spring}:
\begin{equation}
H(q,p) = \frac{p^2}{2m} + \frac{k q^2}{2},
\end{equation}
We generated 50 distinct trajectories by numerically integrating the system using the RK-45 method over the time interval $[0,3]$, with timescale $T=10$. Initial conditions were sampled randomly. The resulting dataset consists of state pairs $(q,p)$ as inputs and their corresponding time derivatives $(\dot{q},\dot{p})$ as targets. An 80/20 train-test split was used, resulting in 1200 training samples and 300 testing samples.
We trained a four-layer HNN with 256 units per hidden layer. The input layer consists of two units corresponding to the generalized coordinate $q$ and momentum $p$, and the network outputs a single scalar Hamiltonian value $H$. The model was trained on a CPU for 100 epochs using a batch size of 64.

\subsubsection{\textbf{Lagrangian Neural Networks (LNN)}}
The Lagrangian of the mass-spring system is given by~\citep{mitofsky_mass_spring}:
\begin{equation}
L(q, \dot{q}) = \frac{1}{2} m \dot{q}^2 - \frac{1}{2} k q^2,    
\end{equation}

We generated 10 samples of trajectory by numerically integrating the system using SciPy’s \texttt{odeint} solver over the time interval $[0,10]$. Initial conditions were sampled uniformly, with $q \in [-2,2]$ and $\dot{q} \in [-1,1]$. The dataset consists of input states $(q,\dot{q})$ and corresponding acceleration values $\ddot{q}$. An 80/20 train–test split was used, resulting in 12{,}000 training samples and 3{,}000 testing samples.
We trained a four-layer LNN with 256 units in each hidden layer. The input layer consists of two units corresponding to $q$ and $\dot{q}$, and the network outputs a scalar Lagrangian value $L$. The model was trained on a CPU for 100 epochs using a batch size of 64.

\subsubsection{\textbf{Symplectic-RNN}}
To generate the dataset for the SRNN model, we created 100 trajectories using a fixed time step of $\Delta t = 0.01$ over a timescale of $T=10$. The dataset consists of sequences of state variables $(q,p)$ used to learn the underlying Hamiltonian dynamics.
We trained a four-layer H-NET model with 256 units per layer. The input layer consists of two units corresponding to the generalized coordinate $q$ and momentum $p$, and the network outputs a scalar Hamiltonian value $H$. The model was trained for 1000 epochs using a learning rate of $10^{-3}$ and a batch size of 32.

\begin{figure}[H]
  \centering
  \includegraphics[width=\linewidth]{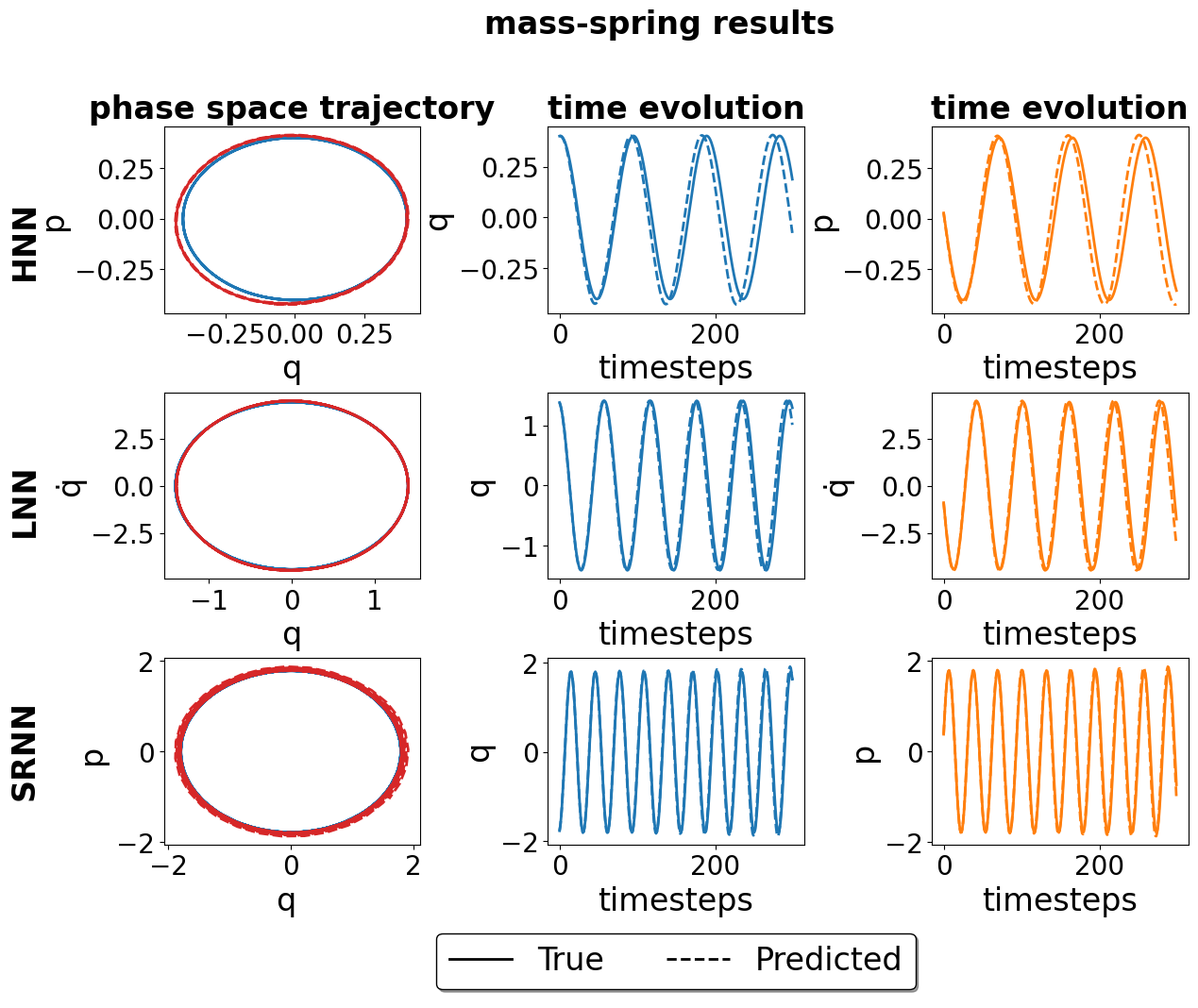}
  \caption{Mass-spring results. First column shows comparison of phase-space trajectories, second column shows comparison of $q$ coordinates, third column shows comparison of $p$ for HNN, SRNN and $\dot{q}$ for LNN.}
  \label{fig:mass_spring_results}
\end{figure}
\begin{table}[H]
\centering
\begin{tabular}{|l|c|c|c|c|c|}
\hline
Model & MSE & MAE & RMSE & STD & VAR \\
\hline
HNN & 0.0120 & 0.0872 & 0.0966 & 0.1091 & 0.0119 \\
LNN & 0.2637 & 0.3553 & 0.4524 & 0.5134 & 0.2636 \\
SRNN & 0.0195 & 0.1106 & 0.1399 & 0.1399 & 0.0195 \\
\hline
\end{tabular}
\caption{Quantitative performance comparison of HNN, LNN, and SRNN on the mass-spring system. Reported metrics are computed between predicted and ground-truth trajectories.}
\label{tab:mass_spring_metrics}
\end{table}
Table~\ref{tab:mass_spring_metrics} summarizes the quantitative performance of all three models on the mass-spring system. Among the compared models, the HNN achieves the lowest error across all metrics, indicating the most accurate recovery of system dynamics. The SRNN shows slightly higher errors than HNN but remains competitive, particularly in long-term trajectory prediction. In contrast, the LNN exhibits the highest numerical error across all metrics. However, despite its larger quantitative errors, the LNN produces qualitatively reasonable trajectories, suggesting that it captures the overall system behavior even though its state-derivative estimates are less accurate. Overall, HNN demonstrates the best numerical accuracy for this system, while SRNN and LNN remain competitive in preserving qualitative dynamical behavior (see Figure \ref{fig:mass_spring_results}).

\subsection{Simple-Pendulum System}
A simple pendulum consists of a point mass $m$ attached to a rod of length $l$, swinging under the influence of gravity. It exhibits periodic motion in a single plane, making it a canonical example of nonlinear oscillatory dynamics.
\subsubsection{\textbf{Hamiltonian Neural Networks (HNN) :}}
The hamiltonian equation is given by~\citep{greydanus_hnn_code}:
\begin{equation}
H =\frac{p^2}{2 m l^2} + m g l \left(1 - \cos q\right)
\end{equation}
We generated 50 trajectories by numerically integrating the system over the time interval $[0,10]$ with a temporal resolution of $T=15$. Initial conditions were sampled randomly. An 80/20 train–test split was used, resulting in 6{,}000 training samples and 1{,}500 testing samples.
We trained a five-layer HNN with 128 units per hidden layer. The input layer consists of two units corresponding to the generalized coordinate $q$ and momentum $p$, and the network outputs a single scalar Hamiltonian value $H$. The model was trained for 100 epochs using a batch size of 64.

\subsubsection{\textbf{Lagrangian Neural Networks :}}
The Lagrangian of simple pendulum is given by~\citep{owenpendulum}:
\begin{equation}
L(\theta, \dot{\theta}) =
\frac{1}{2} m l^{2} \dot{\theta}^{2}
- m g l \left( 1 - \cos \theta \right)
\end{equation}
We generated 10 trajectories by sampling initial conditions uniformly, with angular position $q \in [-\pi, \pi]$ and angular velocity $\dot{q} \in [-2, 2]$. Each trajectory was simulated over the time interval $[0, 10]$. The resulting data were split into training and testing sets using an 80/20 ratio, yielding 800 training samples and 200 testing samples.
We trained a four-layer LNN with 256 units per hidden layer. The input layer consists of two units corresponding to $q$ and $\dot{q}$, and the network outputs a scalar Lagrangian value $L$. The model was trained for 200 epochs using a batch size of 64.

\subsubsection{\textbf{Symplectic-RNN :}}
We generated a training dataset with 100 trajectories by sampling initial conditions randomly for $q$ and $p$ over a time step of $\Delta t = 0.1$ and timescale $T = 10$. A separate test dataset was created using 50 trajectories.
We trained a four-layer H-NET model with 256 units per hidden layer. The input layer consists of two units corresponding to $q$ and $p$, and the network outputs a scalar Hamiltonian value $H$. Training was performed for 1000 epochs with a learning rate of $10^{-3}$ and a batch size of 32.

\begin{figure}[H]
  \centering
  \includegraphics[width=\linewidth]{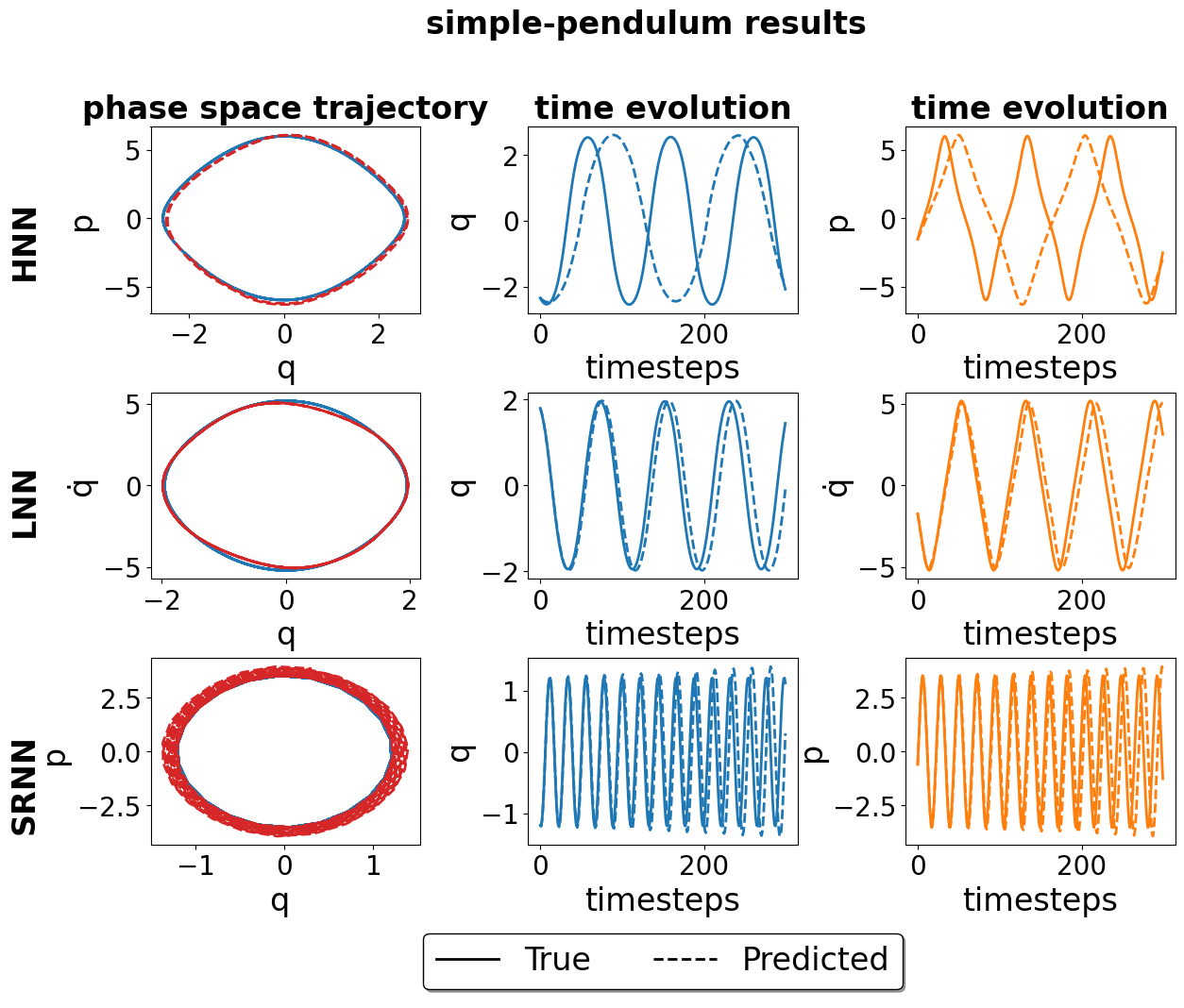}
  \caption{Simple-pendulum results. First column shows comparison of phase-space trajectories, second column shows comparison of $q$ coordinates, third column shows comparison of $p$ for HNN, SRNN and $\dot{q}$ for LNN.}
  \label{fig:simple_pendulum_results}
\end{figure}
\begin{table}[H]
\centering
\begin{tabular}{|l|c|c|c|c|c|}
\hline
Model & MSE & MAE & RMSE & STD & VAR \\
\hline
HNN & 16.859 & 3.1667 & 3.6086 & 4.1060 & 16.8593 \\
LNN & 1.3993 & 0.8749 & 1.1024 & 1.1784 & 1.3886 \\
SRNN & 2.2150 & 0.9814 & 1.4883 & 1.4881 & 2.2145 \\
\hline
\end{tabular}
\caption{Quantitative performance comparison of HNN, LNN, and SRNN on the simple pendulum. Metrics are computed between predicted and ground-truth trajectories.}
\label{tab:simple_pendulum_metrics}
\end{table}
Table~\ref{tab:simple_pendulum_metrics} summarizes the quantitative performance of the three models on the simple pendulum. The LNN achieves the lowest error across all metrics, indicating highly accurate recovery of the system dynamics. The SRNN performs moderately well, outperforming HNN in terms of RMSE, standard deviation, and variance. In contrast, the HNN exhibits a much higher MSE despite qualitatively capturing the phase-space structure of the trajectories, suggesting that it preserves the Hamiltonian geometry even when state-wise prediction errors are large. Overall, while all three models produce qualitatively reasonable trajectories (see Figure \ref{fig:simple_pendulum_results}), LNN demonstrates the best quantitative performance for this system.

\subsection{Spring-pendulum System}
The spring-pendulum system has a mass attached to a spring hanging from a fixed pivot. The mass can move up and down along the spring and swing like a pendulum, resulting in coupled two-dimensional motion.

\subsubsection{\textbf{Hamiltonian Neural Networks (HNN) :}}
The Hamiltonian equation is defined as~\citep{venturi_spring_pendulum}:
\begin{equation}
H = \frac{p_r^2}{2 m} + \frac{p_\theta^2}{2 m r^2} + m g r \cos \theta + \frac{1}{2} k (r - l_0)^2
\end{equation}
We generated 5 trajectories for the spring–pendulum system over the time interval $[0,10]$, using 3000 time steps. The initial conditions were fixed at $r = 1.1$, $\theta = 0.5$, $p_r = 0.0$, and $p_\theta = 0.0$. An 80/20 train–test split was applied, resulting in 12{,}000 training samples and 3{,}000 testing samples.
We trained a four-layer HNN with 256 units per hidden layer. The input layer consists of four units corresponding to the state variables $(r, \theta, p_r, p_\theta)$, and the network outputs a single scalar Hamiltonian value $H$. The model was trained for 200 epochs using a batch size of 64.

\subsubsection{\textbf{Lagrangian Neural Networks :}}
The Lagrangian equation is given as~\citep{venturi_spring_pendulum}:
\begin{equation}
L = \frac{1}{2} m \dot r^2 + \frac{1}{2} m r^2 \dot \theta^2 
- \frac{1}{2} k (r - l_0)^2 - m g r \cos \theta
\end{equation}
We generated 5 trajectories for the spring–pendulum system using fixed initial conditions: $r = 1.1$, $\theta = 0.5$, and $\dot{r} = \dot{\theta} = 0$. Each trajectory was simulated over the time interval $[0,10]$ with 1{,}500 time steps. An 80/20 train–test split was applied, resulting in 6{,}000 training samples and 1{,}500 testing samples.
We trained a four-layer LNN with 256 units per hidden layer. The input layer consists of four units corresponding to the state variables $(r, \theta, \dot{r}, \dot{\theta})$, and the network outputs a scalar Lagrangian value $L$. The model was trained for 100 epochs using a batch size of 64.

\subsubsection{\textbf{Symplectic-RNN :}}
We generated 100 trajectories for the spring–pendulum system with a time step of $\Delta t = 0.01$ and a timescale of $T = 20$.  
We trained a four-layer H-NET model with 256 units per hidden layer. The input layer consists of four units corresponding to $(r, \theta, \dot{r}, \dot{\theta})$, and the network outputs a single scalar Hamiltonian value $H$. The model was trained for 1000 epochs with a learning rate of $10^{-3}$ and a batch size of 32.

\begin{figure}[H]
  \centering
  \includegraphics[width=\linewidth]{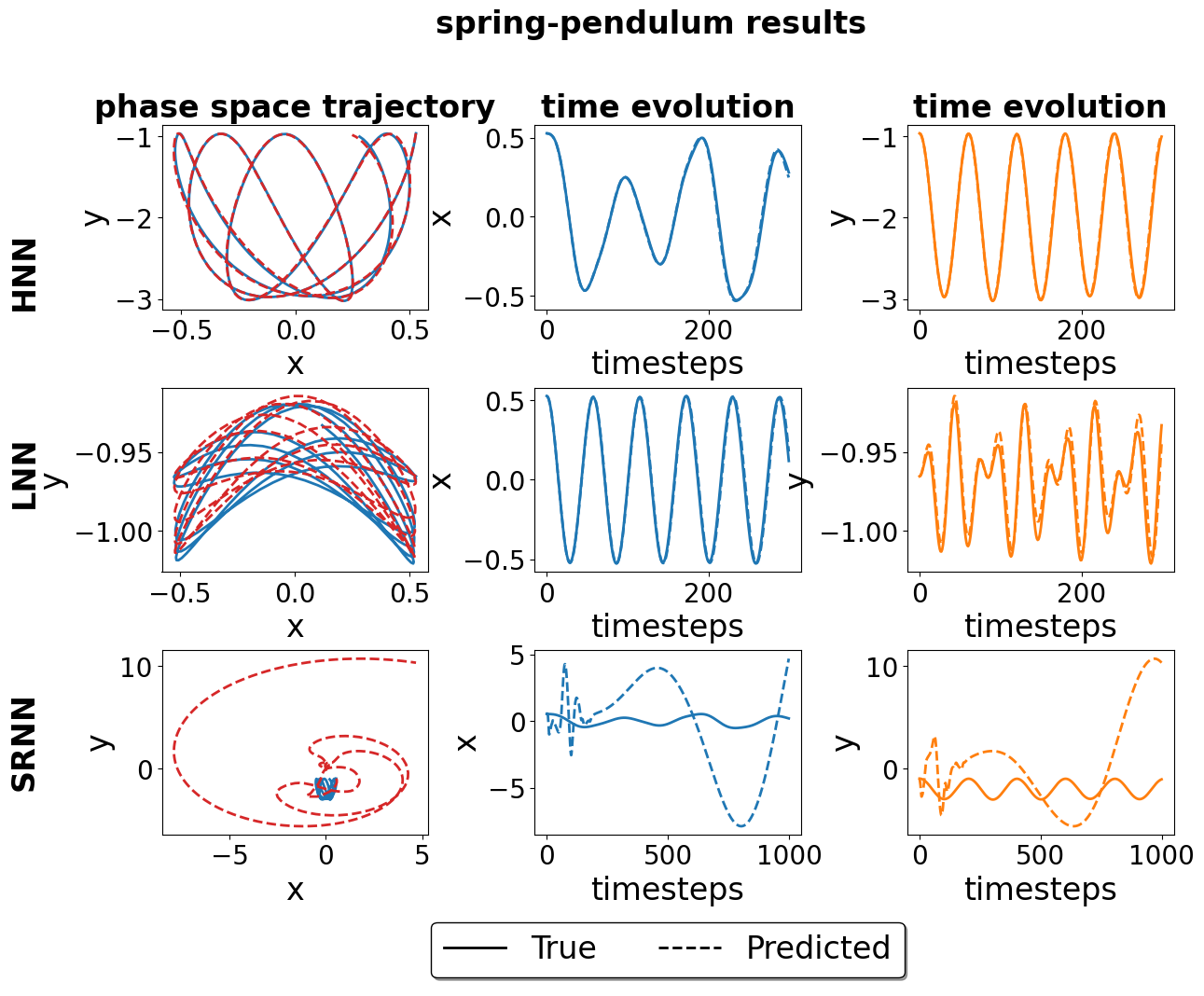}
  \caption{Spring-pendulum results. First column shows comparison of phase-space trajectories, second column shows comparison of $x$ coordinates, third column shows comparison of $y$ coordinates.}
  \label{fig:spring_pendulum_results}
\end{figure}
\begin{table}[H]
\centering
\begin{tabular}{|l|c|c|c|c|c|}
\hline
Model & MSE & MAE & RMSE & STD & VAR \\
\hline
HNN & 0.0008 & 0.0203 & 0.0269 & 0.0294 & 0.0008 \\
LNN & 0.0003 & 0.0129 & 0.0166 & 0.0186 & 0.0003 \\
SRNN & 20.130 & 3.3788 & 4.4208 & 4.3703 & 19.100 \\
\hline
\end{tabular}
\caption{Quantitative performance comparison of HNN, LNN, and SRNN on the spring–pendulum system. Metrics are computed between predicted and ground-truth trajectories.}
\label{tab:spring_pendulum_metrics}
\end{table}
Table~\ref{tab:spring_pendulum_metrics} summarizes the quantitative performance of the three models on the spring–pendulum system. The LNN achieves the lowest error across all metrics, demonstrating highly accurate modeling of the system dynamics. The HNN performs moderately well, showing reasonable accuracy in state predictions and qualitatively preserving the phase-space structure despite slightly higher errors (see Figure \ref{fig:spring_pendulum_results}). The SRNN exhibits the highest errors, likely because it was trained on smaller timesteps, which limited its ability to capture the more chaotic dynamics of the system.

\subsection{Double Pendulum System}
The double pendulum consists of two point masses connected by rigid rods, exhibiting highly nonlinear and chaotic motion due to the coupling between the two pendulums.

\subsubsection{\textbf{Hamiltonian Neural Networks :}}
The Hamiltonian equation is defined as~\citep{baden_double_pendulum} :
\begin{equation*}
H(q_1, q_2, p_1, p_2) = T + V
\end{equation*}
\begin{equation}
\begin{aligned}
T &= \frac{ m_2 l_2^2 p_1^2 + (m_1 + m_2) l_1^2 p_2^2 - 2 m_2 l_1 l_2 p_1 p_2 \cos(q_1 - q_2) }{ 2 \, l_1^2 l_2^2 \left( m_1 + m_2 \sin^2(q_1 - q_2) \right) } \\
V &= - (m_1 + m_2) g l_1 \cos q_1 - m_2 g l_2 \cos q_2
\end{aligned}
\end{equation}
We generated 5 trajectories for the double pendulum system over the time interval $[0,30]$, using 200 time steps. An 80/20 train–test split was applied, resulting in 900 training samples and 100 testing samples.
We trained a four-layer HNN with 256 neurons per hidden layer. The input layer consists of four units corresponding to the generalized coordinates and momenta of the system, and the network outputs a single scalar Hamiltonian value $H$. The model was trained for 1000 epochs using a batch size of 64.

\subsubsection{\textbf{Lagrangian Neural Networks (LNN):}}
The Lagrangian is defined as ~\citep{cranmer_lnn_code}:

\begin{equation*}
L(\theta_1, \theta_2, \dot{\theta}_1, \dot{\theta}_2) = T - V,
\end{equation*}
\begin{equation}
\begin{aligned}
T &= \frac{1}{2} m_1 (l_1 \dot{\theta}_1)^2 
    + \frac{1}{2} m_2 \Big[ (l_1 \dot{\theta}_1)^2 + (l_2 \dot{\theta}_2)^2 
     + 2 l_1 l_2 \dot{\theta}_1 \dot{\theta}_2 \cos(\theta_1 - \theta_2) \Big], \\
V &= m_1 g (-l_1 \cos \theta_1) + m_2 g \big( -l_1 \cos \theta_1 - l_2 \cos \theta_2 \big)
\end{aligned}
\end{equation}

We generated 10 trajectories for the double pendulum system using randomly initialized input conditions. An 80/20 train–test split was applied, resulting in 8{,}000 training samples and 2{,}000 testing samples.
We trained a four-layer LNN with 256 neurons per hidden layer. The input layer consists of four units corresponding to the generalized coordinates and momenta of the system, and the network outputs a single scalar Lagrangian value $L$. The model was trained for 1000 epochs using a batch size of 64.

\subsubsection{\textbf{Symplectic-RNN :}}
We generated 200 trajectories for the double pendulum system with a time step of $\Delta t = 0.2$ and a timescale of $T = 20$.  
We trained a four-layer H-NET model with 128 neurons per hidden layer. The input layer consists of four units corresponding to the generalized coordinates and momenta of the system, and the network outputs a single scalar Hamiltonian value $H$. Training was performed for 200 epochs with a learning rate of $10^{-3}$ and a batch size of 64.

\begin{figure}[H]
  \centering
  \includegraphics[width=\linewidth]{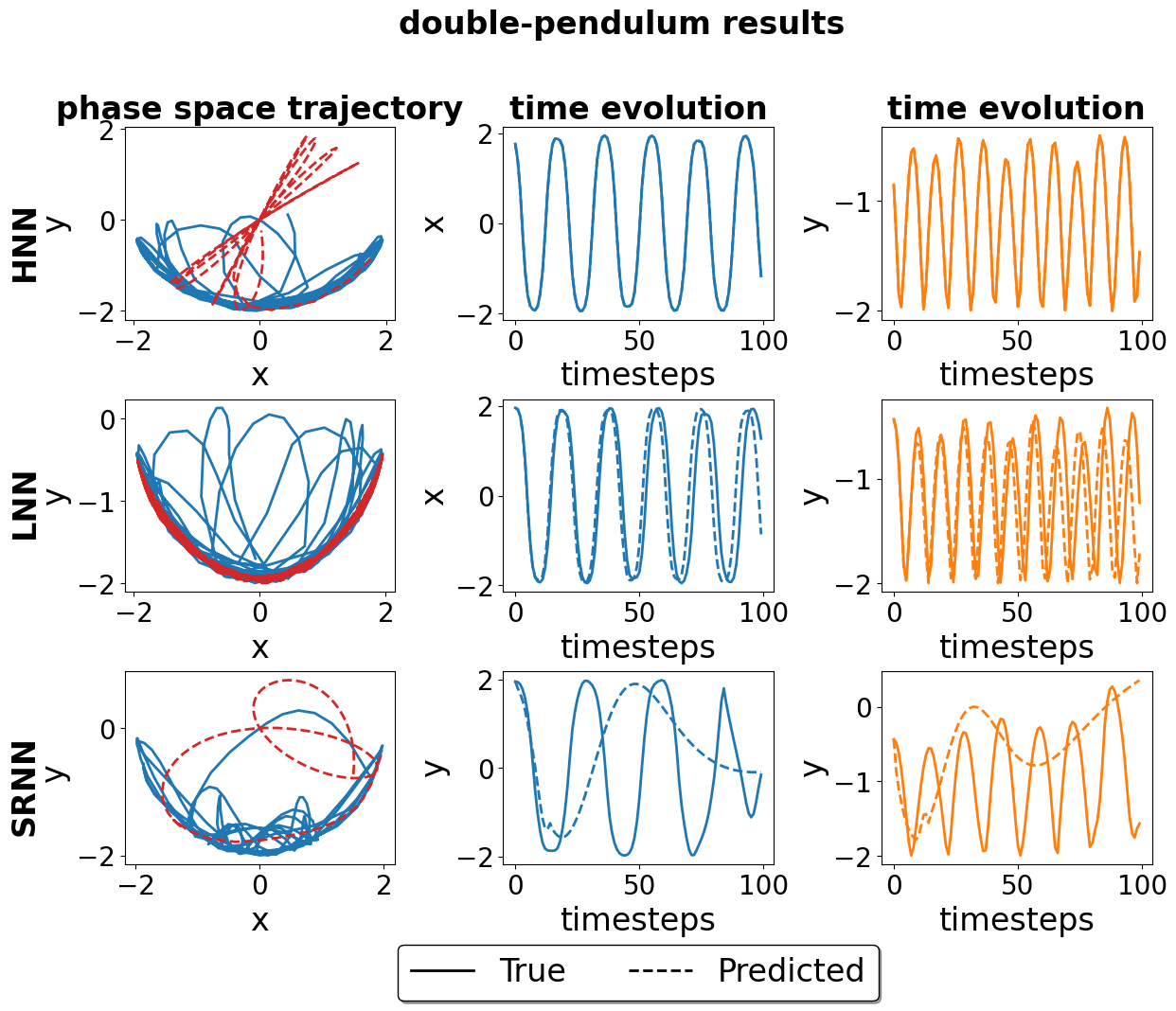}
  \caption{Double-pendulum results. First column shows comparison of phase-space trajectories, second column shows comparison of $x$ coordinates, third column shows comparison of $y$ coordinates}
  \label{fig:double_pendulum_results}
\end{figure}
\begin{table}[H]
\centering
\begin{tabular}{|l|c|c|c|c|c|}
\hline
Model & MSE & MAE & RMSE & STD & VAR \\
\hline
HNN &  1.5451 & 0.9803 & 1.2110 & 1.1911 & 1.4188 \\
LNN & 0.6125 & 0.5579 & 0.7246 & 0.7823 & 0.6121 \\
SRNN & 1.7197 & 1.0758 & 1.2920 & 1.1384 & 1.2959 \\
\hline
\end{tabular}
\caption{Quantitative performance comparison of HNN, LNN, and SRNN on the double–pendulum system. Metrics are computed between predicted and ground-truth trajectories.}
\label{tab:double_pendulum_metrics}
\end{table}
Table~\ref{tab:double_pendulum_metrics} summarizes the quantitative performance of all models on the double pendulum system. Due to the strongly chaotic nature of the double-pendulum, both HNN and SRNN exhibit relatively large prediction errors, as small deviations quickly amplify over time and lead to divergence from the ground-truth trajectories. The LNN achieves significantly lower error across all metrics, indicating more accurate short-term predictions. However, despite its improved numerical performance, the LNN still struggles to fully capture the chaotic phase-space structure over longer time horizons, with trajectories eventually diverging from the ground truth (see Figure \ref{fig:double_pendulum_results}).

\subsection{Bouncing Ball System}
The bouncing ball system models a point mass moving under gravity and undergoing inelastic collisions with a fixed surface, introducing contact dynamics into the system.

\subsubsection{\textbf{Hamiltonian Neural Networks :}}
We model the bouncing ball as a simple particle moving under constant gravity. To simulate the ball's movement, we used the Forward Euler method. At each time step, the model calculates the change in position and momentum using the gradients of the Hamiltonian. To handle the bounces, we added a ground condition: whenever the height $q$ reaches zero, the momentum $p$ is flipped and reduced based on the coefficient of restitution $e$. The hamiltonian is given as ~\citep{wheeler_free_fall} ~\citep{stoyanov_bouncing_ball}:
\begin{equation}
    H(q, p) = \frac{p^2}{2m} + mgq
\label{eq:bouncing_ball_hamiltonian}
\end{equation}
The dataset contains 50 trajectories with initial heights randomly sampled in the range $[2.5, 3.0]$. Each trajectory was simulated over the time interval $[0, 5]$ with a timescale of 100. To account for collisions with the ground, the coefficient of restitution $\rho = 0.8$ was applied during trajectory integration.  

We trained a five-layer HNN with 256 neurons in each hidden layer. The input layer consists of two units corresponding to the vertical position $q$ and momentum $p$, and the network outputs a single scalar Hamiltonian value $H$. The model was trained for 1{,}000 epochs with a batch size of 64. During trajectory integration after training, the same coefficient of restitution $\rho=0.9$ was applied to model the bounces accurately.

\subsubsection{\textbf{Lagrangian Neural networks :}}
We apply the same collision handling mechanism as in the Hamiltonian formulation, where the coefficient of restitution $e$ governs the momentum reversal upon impact with the ground. The system dynamics are derived from the Lagrangian framework, where the Euler-Lagrange equations determine the equations of motion. The Lagrangian is given by~\citep{wheeler_free_fall} ~\citep{stoyanov_bouncing_ball}:
\begin{equation}
L(q, \dot{q}) = \frac{1}{2} m \dot{q}^2 - m g q
\end{equation}
The dataset contains 10 trajectories with initial heights randomly sampled in the range $[2.5, 3.0]$. Each trajectory was simulated over the time interval $[0, 5]$ with a timescale of 100. To account for collisions with the ground, the coefficient of restitution $\rho = 0.8$ was applied during trajectory integration.  
We trained a four-layer HNN with 128 neurons in each hidden layer. The input layer consists of two units corresponding to the vertical position $q$ and velocity $\dot{q}$, and the network outputs a single scalar Hamiltonian value $H$. The model was trained for 500 epochs with a batch size of 64. During trajectory integration after training, the same coefficient of restitution $\rho=0.9$ was applied to model the bounces accurately.

\subsubsection{\textbf{Symplectic-RNN :}}
The training dataset consists of 100 trajectories with a time step $\Delta t = 0.01$, timescale $T = 20$, and initial heights sampled uniformly in the range $[0.05, 0.15]$. The test dataset contains 5 trajectories with $T = 100$ and $\Delta t = 0.01$. The system dynamics are defined using the Hamiltonian in Eq.~\eqref{eq:bouncing_ball_hamiltonian}.
We trained a four-layer H-NET model with 256 neurons in each hidden layer. The model was trained for 1000 epochs using a learning rate of $0.001$ and a batch size of 64. 

\begin{figure}[H]
  \centering
  \includegraphics[width=\linewidth]{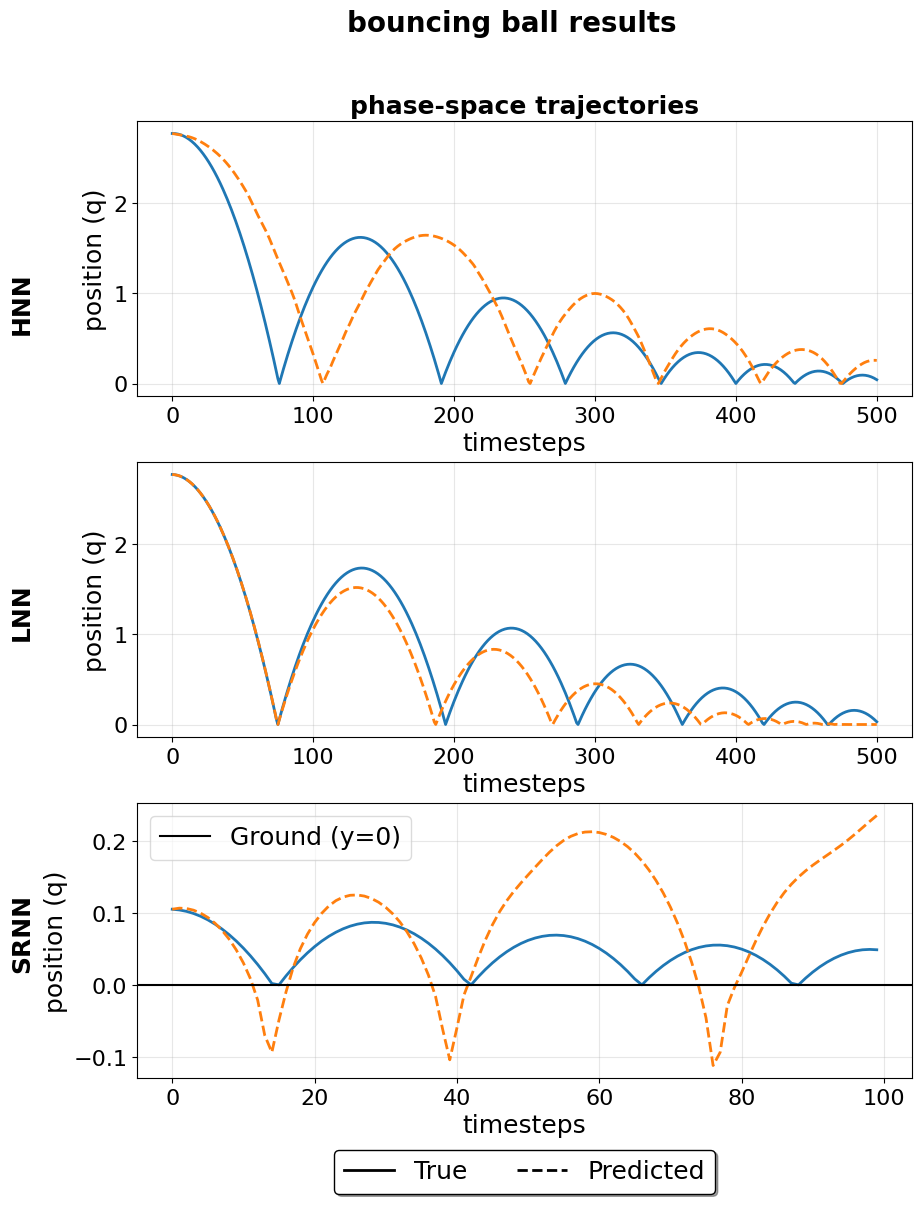}
  \caption{Bouncing ball: comparison of phase-space trajectories of HNN, LNN and SRNN model.}
  \label{fig:bouncing_ball_results}
\end{figure}

\begin{table}[H]
\centering
\begin{tabular}{|l|c|c|c|c|c|}
\hline
Model & MSE & MAE & RMSE & STD & VAR \\
\hline
HNN &  0.3435 & 0.4436 & 0.5861 & 0.5550 & 0.3081 \\
LNN & 0.0599 & 0.1863 & 0.24480 & 0.1934 & 0.0374 \\
SRNN & 0.4463 & 0.3989 & 0.6680 & 0.6694 & 0.4448 \\
\hline
\end{tabular}
\caption{Quantitative performance comparison of HNN, LNN, and SRNN on the bouncing ball system. Metrics are computed between predicted and ground-truth trajectories.}
\label{tab:bouncing_ball_metrics}
\end{table}
As shown in Table~\ref{tab:bouncing_ball_metrics}, the LNN achieves the lowest numerical errors across all metrics, followed by the HNN, while the SRNN exhibits the highest errors. However, despite its higher quantitative error, the HNN appears qualitatively better in the phase-space plots (see Figure \ref{fig:bouncing_ball_results}). The SRNN is trained on trajectories initialized in a much smaller height range $[0.05, 0.15]$, with fewer effective time steps and discrete trajectory updates. As a result, while the HNN and LNN achieves the best quantitative accuracy and the SRNN performs well within its training distribution, the HNN generalizes better to larger-scale phase-space behavior, leading to visually smoother and more physically consistent trajectories.

\subsection{Three-body problem}
The three-body problem is a classical dynamical system that describes the motion of three interacting particles under mutual gravitational forces. It exhibits strong nonlinearity and chaotic behavior, making it a challenging benchmark for evaluating the stability and long-term predictive capability of physics-informed neural networks.

\subsubsection{\textbf{Hamiltonian Neural Networks (HNN) :}}
The Hamiltonian equation for three-body is given as~\citep{greydanus_hnn_code}:
\begin{equation}
H
 =
\sum_{i=1}^{3}
\frac{\|\mathbf{p}_i\|^2}{2 m_i}
\;-\;
\sum_{1 \le i < j \le 3}
\frac{G\, m_i m_j}{\|\mathbf{q}_i - \mathbf{q}_j\|}
\end{equation}
We generated orbital trajectories for a three-body system by simulating 50 trials over the time interval $[0, 3]$ with a timestep $T = 20$. Initial positions were chosen to ensure orbital motion in phase space. The resulting dataset was split into training and testing sets using an 80/20 ratio, resulting in 800 training samples and 200 testing samples.
We trained a 4-layer HNN model with 256 neurons in each hidden layer. The input layer consists of 12 units, corresponding to the positions and momenta of the three bodies (6 position and 6 momentum variables), and the network outputs a single scalar Hamiltonian value. The model was trained for 100 epochs using a batch size of 64.

\subsubsection{\textbf{Lagrangian Neural Networks (LNN) :}}
For the three-body experiment, the dataset was generated by directly computing the accelerations using analytical gravitational interaction equations. The acceleration values are defined as follows ~\citep{krishnaswami2019threebody}:
\begin{equation}
    \mathbf{a}_1 = \frac{G m_2 (\mathbf{r}_2 - \mathbf{r}_1)}{|\mathbf{r}_2 - \mathbf{r}_1|^3} + \frac{G m_3 (\mathbf{r}_3 - \mathbf{r}_1)}{|\mathbf{r}_3 - \mathbf{r}_1|^3}
\end{equation}
\begin{equation}
    \mathbf{a}_2 = \frac{G m_1 (\mathbf{r}_1 - \mathbf{r}_2)}{|\mathbf{r}_1 - \mathbf{r}_2|^3} + \frac{G m_3 (\mathbf{r}_3 - \mathbf{r}_2)}{|\mathbf{r}_3 - \mathbf{r}_2|^3}
\end{equation}
\begin{equation}
    \mathbf{a}_3 = \frac{G m_1 (\mathbf{r}_1 - \mathbf{r}_3)}{|\mathbf{r}_1 - \mathbf{r}_3|^3} + \frac{G m_2 (\mathbf{r}_2 - \mathbf{r}_3)}{|\mathbf{r}_2 - \mathbf{r}_3|^3}
\end{equation}
The dataset consists of 10 distinct trajectories, each initialized with an 18-dimensional input vector. The inputs correspond to the generalized coordinates and velocities $(q, \dot{q})$ of a three-body system embedded in $\mathbb{R}^3$, while the labels represent the corresponding acceleration values.
We trained a four-layer LNN with 256 neurons in each hidden layer and an 18-dimensional input layer for 500 epochs.

\subsubsection{\textbf{Symplectic-RNN :}}
The dataset consists of 100 trajectories for training, with a time step of $\Delta t = 0.01$ and timescale $T = 10$, and a single trajectory for testing with $T = 200$.  
We trained a four-layer H-NET model with 256 neurons per hidden layer and a 12-dimensional input layer corresponding to the positions and momenta of the three bodies. The model was trained for 1000 epochs using a learning rate of $0.003$ and a batch size of 32.

\begin{figure}[H]
  \centering
  \includegraphics[width=\linewidth]{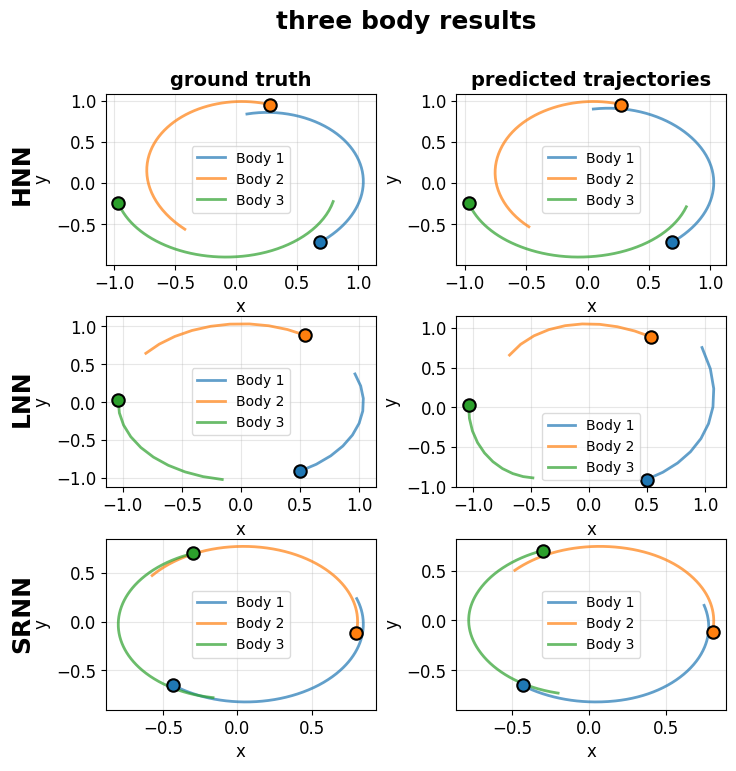}
  \caption{Three-body problem results. Each row shows a different model (HNN, LNN, SRNN). left column shows ground truth, right column shows predictions.}
  \label{fig:3body_results}
\end{figure}
\begin{table}[H]
\centering
\begin{tabular}{|l|c|c|c|c|c|}
\hline
Model & MSE & MAE & RMSE & STD & VAR \\
\hline
HNN &  0.0004 & 0.0156 & 0.0215 & 0.0209 & 0.0004 \\
LNN & 0.0171 & 0.0840 & 0.1086 & 0.1262 & 0.0159 \\
SRNN & 0.0022 & 0.0338 & 0.0479 & 0.0477 & 0.0022 \\
\hline
\end{tabular}
\caption{Quantitative performance comparison of HNN, LNN, and SRNN on the three-body problem. Metrics are computed between predicted and ground-truth trajectories.}
\label{tab:three_body_metrics}
\end{table}
Table~\ref{tab:three_body_metrics} summarizes the quantitative performance of the models on the three-body system. The HNN achieves the lowest error across all metrics, indicating the most accurate trajectory predictions. The SRNN performs competitively, with slightly higher errors but still maintaining stable and accurate orbital dynamics. The LNN exhibits higher numerical errors compared to the other models; however, it still performs well qualitatively, producing stable trajectories and capturing the overall orbital structure of the system (see Figure \ref{fig:3body_results}).

\section{Discussion}
The benchmark results reveal distinct performance patterns across different types of dynamical systems. For stable conservative systems like mass-spring, pendulum, and spring-pendulum, all three architectures demonstrate strong performance both quantitatively and qualitatively, successfully preserving energy conservation and maintaining stable long-term trajectory predictions. The three-body problem with orbital configurations also shows good results due to the relatively simple setup, though three-body systems can exhibit chaotic behavior under different initial conditions. This confirms that physics-informed architectures are well-suited for learning systems with regular dynamics where symplectic structure and conservation laws dominate. However, performance degrades notably for chaotic and dissipative systems. The double pendulum's sensitive dependence on initial conditions and the bouncing ball's discontinuous collision dynamics present greater challenges, limiting long-term prediction accuracy despite reasonable short-term performance.

\section{Conclusion}
We presented a systematic benchmark of three physics-informed neural network architectures such as HNN, LNN, and SRNN implemented within the DeepChem framework across six dynamical systems. Our results demonstrate that these models excel at learning stable conservative systems, achieving strong performance in both quantitative metrics and qualitative trajectory reconstruction.
However, non-conservative and chaotic systems remain challenging for current physics-informed architectures. The double pendulum and bouncing ball experiments reveal fundamental limitations in learning systems with sensitive initial condition. Future work may explore hybrid approaches that combine physics-informed architectures with specialized modules for handling discontinuities, or investigate alternative training strategies that better account for chaotic divergences.

\section*{GenAI Disclosure}
Large language models were used to improve the language, grammar, and clarity of the manuscript. All scientific contributions and analyses are entirely the work of the authors.

\bibliographystyle{ACM-Reference-Format}
\bibliography{references}

@article{greydanus2019hamiltonian,
  title={Hamiltonian neural networks},
  author={Greydanus, Samuel and Dzamba, Misko and Yosinski, Jason},
  journal={Advances in neural information processing systems},
  volume={32},
  year={2019}
}

@article{cranmer2020lagrangian,
  title={Lagrangian neural networks},
  author={Cranmer, Miles and Greydanus, Sam and Hoyer, Stephan and Battaglia, Peter and Spergel, David and Ho, Shirley},
  journal={arXiv preprint arXiv:2003.04630},
  year={2020}
}

@article{chen2019symplectic,
  title={Symplectic recurrent neural networks},
  author={Chen, Zhengdao and Zhang, Jianyu and Arjovsky, Martin and Bottou, L{\'e}on},
  journal={arXiv preprint arXiv:1909.13334},
  year={2019}
}

@phdthesis{ramsundar2018molecular,
  title={Molecular machine learning with DeepChem},
  author={Ramsundar, Bharath},
  year={2018},
  school={Stanford University}
}

@article{lagaris1998artificial,
  title={Artificial neural networks for solving ordinary and partial differential equations},
  author={Lagaris, Isaac E and Likas, Aristidis and Fotiadis, Dimitrios I},
  journal={IEEE transactions on neural networks},
  volume={9},
  number={5},
  pages={987--1000},
  year={1998},
  publisher={IEEE}
}

@article{zhong2021extending,
  title={Extending lagrangian and hamiltonian neural networks with differentiable contact models},
  author={Zhong, Yaofeng Desmond and Dey, Biswadip and Chakraborty, Amit},
  journal={Advances in Neural Information Processing Systems},
  volume={34},
  pages={21910--21922},
  year={2021}
}

@article{singh2025chemberta,
  title={ChemBERTa-3: an open source training framework for chemical foundation models},
  author={Singh, Riya and Barsainyan, Aryan Amit and Irfan, Rida and Amorin, Connor Joseph and He, Stewart and Davis, Tony and Thiagarajan, Arun and Sankaran, Shiva and Chithrananda, Seyone and Ahmad, Walid and others},
  journal={Digital Discovery},
  year={2025},
  publisher={Royal Society of Chemistry}
}

@article{ahmad2022chemberta,
  title={Chemberta-2: Towards chemical foundation models},
  author={Ahmad, Walid and Simon, Elana and Chithrananda, Seyone and Grand, Gabriel and Ramsundar, Bharath},
  journal={arXiv preprint arXiv:2209.01712},
  year={2022}
}

@article{chithrananda2020chemberta,
  title={ChemBERTa: large-scale self-supervised pretraining for molecular property prediction},
  author={Chithrananda, Seyone and Grand, Gabriel and Ramsundar, Bharath},
  journal={arXiv preprint arXiv:2010.09885},
  year={2020}
}

@article{le2024contact,
  title={Contact models in robotics: a comparative analysis},
  author={Le Lidec, Quentin and Jallet, Wilson and Montaut, Louis and Laptev, Ivan and Schmid, Cordelia and Carpentier, Justin},
  journal={IEEE Transactions on Robotics},
  year={2024},
  publisher={IEEE}
}

@article{aljalbout2025reality,
  title={The reality gap in robotics: Challenges, solutions, and best practices},
  author={Aljalbout, Elie and Xing, Jiaxu and Romero, Angel and Akinola, Iretiayo and Garrett, Caelan Reed and Heiden, Eric and Gupta, Abhishek and Hermans, Tucker and Narang, Yashraj and Fox, Dieter and others},
  journal={Annual Review of Control, Robotics, and Autonomous Systems},
  volume={9},
  year={2025},
  publisher={Annual Reviews}
}

@misc{greydanus_hnn_code,
  author       = {Samuel Greydanus},
  title        = {Code for our paper "Hamiltonian Neural Networks"},
  year         = {2019},
  howpublished = {\url{https://github.com/greydanus/hamiltonian-nn}},
  note         = {2026}
}

@misc{cranmer_lnn_code,
    author = {Miles Cranmer},
    title = {Lagrangian Neural Networks},
    year = {2020},
    howpublished = {\url{https://github.com/MilesCranmer/lagrangian_nns}},
    note = {2026}
}

@misc{mitofsky_mass_spring,
    title = {Mass Spring Example},
    author={Andrea M. Mitofsky},
    howpublished={LibreTexts, Section 11.4, CC BY-NC 4.0},
    url={https://eng.libretexts.org/Bookshelves/Electrical_Engineering/Electro-Optics/Direct_Energy_(Mitofsky)/11%3A_Calculus_of_Variations/11.04%3A_Mass_Spring_Example},
    note = {2026}
}

@misc{owenpendulum,
  author       = {Frank Owen},
  title        = {Simple Pendulum via Lagrangian Mechanics},
  howpublished = {Alpha Omega Engineering, Inc.},
  url          = {http://www.aoengr.com/Dynamics/LagrangianMechanicsPendulum.pdf},
  note         = {2026}
}

@misc{venturi_spring_pendulum,
  author       = {Daniele Venturi},
  title        = {Lagrangian and Hamiltonian Dynamics},
  howpublished = {Course notes, AM 224, University of California, Santa Cruz},
  url          = {https://venturi.soe.ucsc.edu/sites/default/files/COURSE_NOTE_8_Lagrangian_and_Hamiltonian_dynamics.pdf},
  note         = {2026}
}

@misc{baden_double_pendulum,
  author       = {Drew Baden},
  title        = {Pendulum Numerical Integration},
  year         = {2019},
  howpublished = {\url{https://physics.umd.edu/hep/drew/numerical_integration/pendulum2.html}},
  note         = {2026}
}

@misc{wheeler_free_fall,
  author       = {Nicholas Wheeler},
  title        = {Classical and Quantum Dynamics in a Uniform Gravitational Field: A. Unobstructed Free Fall},
  howpublished = {Reed College Physics Department},
  year         = {2002},
  url          = {https://www.reed.edu/physics/faculty/wheeler/documents/Quantum%20Mechanics/Miscellaneous%20Essays/Quantum%20Bouncer/E4.%20Free%20Fall.pdf},
  note         = {2026}
}

@misc{stoyanov_bouncing_ball,
  author = {Georgi Z. Stoyanov},
  title  = {The Bouncing Ball},
  year   = {2019},
  howpublished = {Kaggle Notebook},
  url = {https://www.kaggle.com/code/gzstoyanov/the-bouncing-ball},
  note = {2026}
}

@article{krishnaswami2019threebody,
  author  = {Govind S. Krishnaswami and Himalaya Senapati},
  title   = {An Introduction to the Classical Three-Body Problem: From Periodic Solutions to Instabilities and Chaos},
  journal = {Resonance},
  url     = {https://www.ias.ac.in/article/fulltext/reso/024/01/0087-0114},
  year    = {2019}
}

@ARTICLE{2020SciPy-NMeth,
  author  = {Virtanen, Pauli and Gommers, Ralf and Oliphant, Travis E. and
            Haberland, Matt and Reddy, Tyler and Cournapeau, David and
            Burovski, Evgeni and Peterson, Pearu and Weckesser, Warren and
            Bright, Jonathan and {van der Walt}, St{\'e}fan J. and
            Brett, Matthew and Wilson, Joshua and Millman, K. Jarrod and
            Mayorov, Nikolay and Nelson, Andrew R. J. and Jones, Eric and
            Kern, Robert and Larson, Eric and Carey, C J and
            Polat, {\.I}lhan and Feng, Yu and Moore, Eric W. and
            {VanderPlas}, Jake and Laxalde, Denis and Perktold, Josef and
            Cimrman, Robert and Henriksen, Ian and Quintero, E. A. and
            Harris, Charles R. and Archibald, Anne M. and
            Ribeiro, Ant{\^o}nio H. and Pedregosa, Fabian and
            {van Mulbregt}, Paul and {SciPy 1.0 Contributors}},
  title   = {{{SciPy} 1.0: Fundamental Algorithms for Scientific
            Computing in Python}},
  journal = {Nature Methods},
  year    = {2020},
  volume  = {17},
  pages   = {261--272},
  adsurl  = {https://rdcu.be/b08Wh},
  doi     = {10.1038/s41592-019-0686-2},
}

\clearpage
\appendix
\section{Appendix}

\subsection{Preliminary Concepts}

\subsubsection{Deepchem Framework for model training}

DeepChem~\citep{ramsundar2018molecular} provides high-level abstractions such as \texttt{NumpyDataset} and \texttt{TorchModel} that streamline the development and evaluation of machine learning models for scientific applications. While originally designed for molecular modeling and drug discovery, its modular architecture makes it well-suited for implementing and benchmarking physics-informed deep learning models. 

\textbf{NumpyDataset} is DeepChem's in-memory dataset implementation that stores data using NumPy arrays. It's one of the core building blocks in DeepChem and provides a common interface for handling machine learning data. This makes it easy to work with data from different sources and integrate with various machine learning frameworks. NumpyDataset handles standard operations like batch iteration, data transformations.

\textbf{TorchModel} gives us a common interface for working with PyTorch models through methods like \texttt{.fit()},  \texttt{.predict()}. It takes care of the training loop, gradient computation, loss tracking, and saving checkpoints, so we can concentrate on designing the model architectures themselves.

\subsubsection{\textbf{Hamiltonian Systems}}
Hamiltonian is considered as total energy of a system which is sum of Kinetic energy $T$ and Potential energy $V$:
\begin{equation}
H = T + V     
\end{equation}
The Hamiltonian neural networks (HNNs) uses parameterized Neural networks which outputs scalar $H$ value. The input to the network contains state vector (q, p) values simulated over time steps, $q = (q_1, q_2, ...q_n)$ represents position and $p = (p_1, p_2, ...p_n)$ represents momentum variable.
\begin{equation}
\frac{dq}{dt} = \frac{\partial H}{\partial p}, \quad
\frac{dp}{dt} = -\frac{\partial H}{\partial q}
\label{eq:symplectic_gradient}
\end{equation}
Next, we compute the gradients of scalar hamiltonian values $H = (q, p)$ to get $\dot{q}$ and $\dot{p}$. the loss is get calculated by comparing the predicted $\dot{q}$ and $\dot{p}$ with observed values over the period of time $t$:
\begin{equation}
\mathcal{L}_{HNN} = \left\| \frac{\partial H_\theta}{\partial \mathbf{p}} - \frac{\partial \mathbf{q}}{\partial t} \right\|^2 + \left\| \frac{\partial H_\theta}{\partial \mathbf{q}} + \frac{\partial \mathbf{p}}{\partial t} \right\|^2
\end{equation}
After training the model, a preferred ODEsolver (generally RK45 ) is used to integrate the model over the desired timescale which generate the trajectories:

\begin{equation}
    trajectories = ODEsolver(f(q, p), \Delta t, T)
\end{equation}
This allows to test the model on time evolution of position and momenta to ensure whether model performs consistently and preserves the energy.

\subsubsection{\textbf{Lagrangian Systems}}
Lagrangian is defined as difference between kinetic energy $T$ and potential energy $V$:
\begin{equation}
L = T - V     
\end{equation}
Lagrangian Neural Networks (LNNs) use parameterized neural networks which outputs scalar $L$ value. The input to the network contains state vector $(q, \dot{q})$ values simulated over time steps, $q = (q_1, q_2, ...q_n)$ represents position and $\dot{q} = (\dot{q}_1, \dot{q}_2, ...\dot{q}_n)$ represents velocity variable.
\begin{equation}
\frac{d}{dt}\left(\frac{\partial L}{\partial \dot{q}}\right) - \frac{\partial L}{\partial q} = 0
\end{equation}
we can also write the vectorized form of above equation as:
\begin{equation}
\frac{d}{dt}\nabla_{\dot{q}} L = \nabla_q L
\end{equation}
where $(\nabla_{\dot{q}})_i \equiv \frac{\partial}{\partial \dot{q}_i}$, The chain rule is used to expand the time derivative $\frac{d}{dt}$ which results into two new terms with $\ddot{q}$ and $\dot{q}$:
\begin{equation}
(\nabla_{\dot{q}} \nabla_{\dot{q}}^{\top} L)\ddot{q} + (\nabla_q \nabla_{\dot{q}}^{\top} L)\dot{q} = \nabla_q L
\end{equation}
\begin{equation}
\ddot{q} = (\nabla_{\dot{q}} \nabla_{\dot{q}}^{\top} L)^{-1} [\nabla_q L - (\nabla_q \nabla_{\dot{q}}^{\top} L)\dot{q}]
\end{equation}

The loss is calculated by comparing the predicted $\ddot{q}_L$ with observed acceleration $\ddot{q}_{true}$ over the period of time $t$:
\begin{equation}
\mathcal{L}_{LNN} = \left\| \ddot{q}_L - \ddot{q}_{true} \right\|^2
\end{equation}
After training the model, a preferred ODEsolver (generally RK45) is used to integrate the model over the desired timescale which generate the trajectories:
\begin{equation}
    trajectories = ODEsolver(f(q, \dot{q}), \Delta t, T)
\end{equation}
This enables validation of the learned Lagrangian dynamics and verification of energy conservation properties with acceleration values.

\subsubsection{\textbf{Symplectic-RNN}}
Symplectic Recurrent Neural Networks (SRNNs) extends Hamiltonian Neural networks (HNNs) by explicitly modeling time evolution through symplectic integrators. The hamiltonian is defined as $H(q, p) = K(p) + V(q)$ where $K$ is kinetic energy and $V$ is potential energy. Hence the Symplectic gradients equation (eq.~\ref{eq:symplectic_gradient}) gets further simplified as:
\begin{equation}
    \dot{p} = -V'(q), \ \dot{q} = K'(p)
\end{equation}
To integrate the system dynamics, SRNNs uses symplectic integrators most commonly leapfrog method,which is a second order symplectic integrator. Unlike ODE solvers where small numeric errors at each timesteps gets accumulate which leads to drift in energy and instability (Lambert et al), the leapfrog integrator preserves underlying symplectic structure of hamiltonian systems (McLachlan et al).
\begin{equation}
\begin{aligned}
p_{n+1/2} &= p_n - \tfrac{1}{2}\,\Delta t\,V'(q_n), \\
q_{n+1} &= q_n + \Delta t\,K'(p_{n+1/2}), \\
p_{n+1} &= p_{n+1/2} - \tfrac{1}{2}\,\Delta t\,V'(q_{n+1})
\end{aligned}
\end{equation}
During training, the SRNN learns a parametric function $f_\theta(z, t)$, where $z$ represents the system state and $\theta$ denotes the network parameters. Given an observed trajectory $\{z_i\}_{i=0}^{T}$ measured at uniformly spaced time points $\{t_i\}_{i=0}^{T}$, the model generates a predicted trajectory using the leapfrog integrator as shown in Eq.~\ref{eq:integrator}:
\begin{equation}
\label{eq:integrator}
    \left\{ \hat{z}_i(\theta) \right\}_{i=0}^{T}
= \operatorname{Integrator}\!\left( z_0, f_\theta, \{ t_i \}_{i=0}^{T} \right)
\end{equation}
The model is trained by minimizing the mean squared error between the observed trajectory and the predicted trajectory (Eq.~\ref{eq:loss}):
\begin{equation}
\label{eq:loss}
    \mathcal{L}(\theta) = \sum_{i=1}^{T} \left\lVert z_i - \hat{z}_i(\theta) \right\rVert_2^2
\end{equation}

\subsection{Pseudocodes for models}

This section shows pseudocodes for training and integration of HNN (\ref{algo-hnn}), LNN (\ref{algo-lnn}) and SRNN (\ref{algo-srnn}) models.

\begin{algorithm}[H]
    \caption{Training and Integration of HNN model}
    \label{algo-hnn}
    \begin{algorithmic}[1]
        \STATE \textbf{Training:}
        \STATE \textbf{Dataset:} Input as System state $(q_n, p_n)$ and Labels as time derivative: $(\dot{q}_n, \dot{p}_n)$
        \FOR{each epoch}
            \STATE Forward pass to get the scalar hamiltonian value (total energy) - $H = \text{MLP}(q_n, p_n)$
            \STATE calculate gradients of scalar $H$ with $p$ to get $\dot{q}_{pred} = \frac{\partial H}{\partial p_n}$ and with $q$ to get \quad $\dot{p}_{pred} = -\frac{\partial H}{\partial q_n}$
            \STATE Mean Square Error (MSE) loss between predicted and observed time derivatives values$\mathcal{L} = \|\dot{q}_{pred} - \dot{q}_n\|^2 + \|\dot{p}_{pred} - \dot{p}_n\|^2$
            \STATE Update the HNN parameters by backpropagating the loss $\mathcal{L}$ using gradient descent

        \ENDFOR
        \STATE
        \STATE \textbf{Integration:}
        \STATE \textbf{Input:} Initial state $(q_0, p_0)$, time steps $t$
        \STATE Define dynamics: $f(q, p) = \left(\frac{\partial H}{\partial p}, -\frac{\partial H}{\partial q}\right)$
        \STATE $(q(t), p(t)) = \text{ODESolver}(f, (q_0, p_0), t)$
        \STATE \textbf{Output:} Predicted trajectory $(q(t), p(t))$
    \end{algorithmic}
\end{algorithm}

\begin{algorithm}[H]
    \caption{Training and Integration of LNN model}
    \label{algo-lnn}
    \begin{algorithmic}[1]
        \STATE \textbf{Training:}
        \STATE \textbf{Dataset:} Input as system state $(q_n, \dot{q}_n)$ and labels as accelerations $(\ddot{q}_n)$
        \FOR{each epoch}
            \STATE Forward pass to get scalar Lagrangian value
            \STATE \quad $L = \text{MLP}(q_n, \dot{q}_n)$
            \STATE Compute first-order derivatives (jacobians):
            \STATE \quad $\frac{\partial L}{\partial q_n}, \; \frac{\partial L}{\partial \dot{q}_n}$
            \STATE Compute second-order derivatives (hessians):
            \STATE \quad $\frac{\partial^2 L}{\partial \dot{q}_n^2}, \; \frac{\partial^2 L}{\partial \dot{q}_n \partial q_n}$
            \STATE Solve Euler-Lagrange equation to get accelerations:
            \STATE \quad $\ddot{q}_{pred} =
            \left(\frac{\partial^2 L}{\partial \dot{q}^2}\right)^{-1}
            \left(
            \frac{\partial L}{\partial q}
            -
            \frac{\partial^2 L}{\partial \dot{q} \partial q}\dot{q}
            \right)$
            \STATE Mean Square Error (MSE) loss:
            \STATE \quad $\mathcal{L} = \|\ddot{q}_{pred} - \ddot{q}_n\|^2$
            \STATE Update LNN parameters by backpropagating loss $\mathcal{L}$ using gradient descent
        \ENDFOR
        \STATE
        \STATE \textbf{Integration:}
        \STATE \textbf{Input:} Initial state $(q_0, \dot{q}_0)$, time steps $t$
        \STATE Define dynamics:
        \STATE \quad $f(q, \dot{q}) = (\dot{q}, \ddot{q}_{pred})$
        \STATE $(q(t), \dot{q}(t)) = \text{ODESolver}(f, (q_0, \dot{q}_0), t)$
        \STATE \textbf{Output:} Predicted trajectory $(q(t), \dot{q}(t))$
    \end{algorithmic}
\end{algorithm}

\begin{algorithm}[H]
    \caption{Training and Integration of SRNN model}
    \label{algo-srnn}
    \begin{algorithmic}[1]
        \STATE \textbf{Training:}
        \STATE \textbf{Dataset:} Observed trajectory $\{z_i\}_{i=0}^{T}$ where $z_i = (q_i, p_i)$ at time points $\{t_i\}_{i=0}^{T}$
        \FOR{each epoch}
            \STATE Extract initial state: $z_0 = (q_0, p_0)$
            \STATE Forward pass through H-NET to get scalar Hamiltonian: $H = \text{H-NET}(q, p)$
            \STATE Generate predicted trajectory using leapfrog integrator:
            \STATE \quad $\left\{ \hat{z}_i(\theta) \right\}_{i=0}^{T} = \operatorname{Integrator}\!\left( z_0, f_\theta, \{ t_i \}_{i=0}^{T} \right)$
            \STATE Compute Mean Square Error (MSE) loss: $\mathcal{L}= \sum_{i=1}^{T} \left\| z_i - \hat{z}_i(\theta) \right\|_2^2$
            \STATE Update H-NET parameters $\theta$ by backpropagating the loss $\mathcal{L}(\theta)$ using gradient descent
        \ENDFOR
        \STATE
        \STATE \textbf{Integration (Testing):}
        \STATE \textbf{Input:} Initial state from test dataset $z_0 = (q_0, p_0)$, time points $\{t_i\}_{i=0}^{T}$
        \STATE Forward pass through trained H-NET: $H = \text{H-NET}(q, p)$
        \STATE Generate predicted trajectory: $\left\{ \hat{z}_i(\theta) \right\}_{i=0}^{T} = \operatorname{Integrator}\!\left( z_0, f_\theta, \{ t_i \}_{i=0}^{T} \right)$
        \STATE \textbf{Output:} Predicted trajectory $\{(q_i, p_i)\}_{i=0}^{T}$
    \end{algorithmic}
\end{algorithm}

\end{document}